\def\year{2020}\relax
\pdfoutput=1
\documentclass[letterpaper]{article} % DO NOT CHANGE THIS
\usepackage{aaai20}  % DO NOT CHANGE THIS
\usepackage{times}  % DO NOT CHANGE THIS
\usepackage{helvet} % DO NOT CHANGE THIS
\usepackage{courier}  % DO NOT CHANGE THIS
\usepackage[hyphens]{url}  % DO NOT CHANGE THIS
\usepackage{graphicx} % DO NOT CHANGE THIS
\usepackage{graphicx}  % DO NOT CHANGE THIS
\usepackage{amsmath}

\usepackage{algorithm}% http://ctan.org/pkg/algorithm
\usepackage{algpseudocode}% http://ctan.org/pkg/algorithmicx

\title{OAAE: Adversarial Autoencoders for Novelty Detection in Multi-modal Normality Case via Orthogonalized Latent Space}
\author{Sungkwon An\textsuperscript{\rm 1}\thanks{Authors contributed equally}, \Large \textbf{Jeonghoon Kim}\textsuperscript{\rm 3}\footnotemark[1], \textbf{Myungjoo Kang}\textsuperscript{\rm 2}, \textbf{Shahbaz Razaei}\textsuperscript{\rm 4}, \textbf{Xin Liu}\textsuperscript{\rm 4}\thanks{Corresponding author}\\ \\ % All authors must be in the same font size and format. Use \Large and \textbf to achieve this result when breaking a line
\textsuperscript{\rm 1}Computational Science and Technology, Seoul National University, Korea\\
\textsuperscript{\rm 2}Department of Mathematics, Seoul National University, Korea\\
\{sk\_an, mkang\} @snu.ac.kr\\
\textsuperscript{\rm 3}Department of Mathematics, University of California, Davis, CA\\
\textsuperscript{\rm 4}Department of Computer Science, University of California, Davis, CA\\
\{jhxkim, srezaei, xinliu\}@ucdavis.edu}

\begin{document}

\maketitle

\begin{abstract}
Novelty detection using deep generative models such as autoencoder, generative adversarial networks mostly takes image reconstruction error as novelty score function. However, image data, high dimensional as it is, contains a lot of different features other than class information which makes models hard to detect novelty data. The problem gets harder in multi-modal normality case. To address this challenge, we propose a new way of measuring novelty score in multi-modal normality cases using orthogonalized latent space. Specifically, we employ orthogonal low-rank embedding in the latent space to disentangle the features in the latent space using mutual class information. With the orthogonalized latent space, novelty score is defined by the change of each latent vector. Proposed algorithm was compared to state-of-the-art novelty detection algorithms using GAN such as RaPP and OCGAN, and experimental results show that ours outperforms those algorithms.
\end{abstract}

\section{Introduction}
\begin{figure}[t]
\centering
\includegraphics[width=\columnwidth]{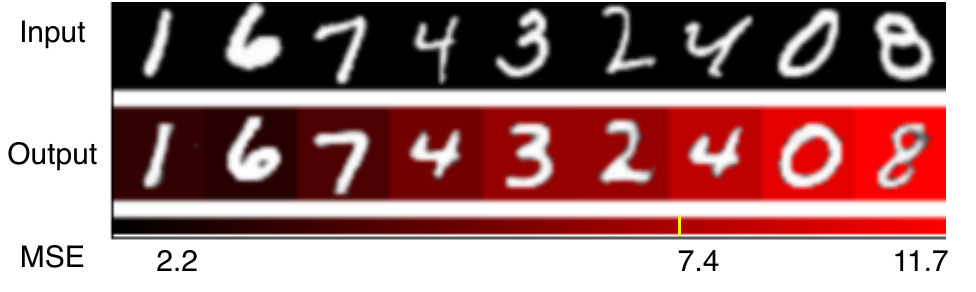}
\caption{Limitation of novelty detection using image reconstruction error. Top: Input images. Middle: Output images of adversarial autoencoder (AAE). Bottom: mean squared error (MSE) of all images. We set the images of digits of 0-8 and 9 as normal and novelty, respectively. Since mean of novelty scores among the image of digit 9 (novelty class) is 7.4, MSE values of normal image bigger than 7.4 lead to wrong novelty detection. }
\label{figwrongrecon}
\end{figure}

Novelty detection, also called anomaly detection in broader perspective, is regarded to be a task of recognising the test data that differs in some respect from the data that are previously seen. Novelty detection has been actively researched since the demand has been increasing due to its significance and broad applications in security, AI safety, healthcare industry.\\
Deep learning has recently shown tremendous performances in learning distribution and representations of various complicated data such as high-dimensional data, time series data. Deep learning for novelty detection aims to learn feature representations and output novelty scores through the neural network to detect data, which has different feature representations from the previously observed data. Many deep learning algorithms for novelty detection has been proposed recently, showing  significantly better performances than traditional novelty detection methods. Deep generative models such as autoencoder (AE), generative adversarial networks (GANs) and their variational models are recognized as one of the biggest breakthrough in deep learning. Since they show great performances in pattern recognition in general, they are adopted for novelty detection in deep learning framework frequently. Deep generative models-based novelty detection algorithms such as OCGAN~\cite{ocgan}, RaPP~\cite{rapp}, AnoGAN~\cite{anogan}, ~\cite{anovae}, and ~\cite{sakurada2014anomaly} usually takes image reconstruction error or extension of it as a novelty score function. The key in novelty detection is to differentiate whether the input data is normal or novelty. However, as image data itself has a lot of inherent traits, e.g. rotations and thickness of the digit in images in MNIST dataset, image reconstruction error can be magnified by those factors, which eventually increases the wrong novelty detection cases potentially as shown in Figure \ref{figwrongrecon}. This gets worse in multi-modal normality case, which we aim to tackle. To the best of our knowledge, there has not been any precedent deep generative approaches to tackle novelty detection in multi-modal normality cases.\\
In this paper, we propose a new framework of novelty score function using orthogonalized latent space. Detection of novelty class in latent space has several benefits. Latent space is lower dimensional space with the feature information than the original high dimensional data, which is easier to be handled. Furthermore, features in latent space can be disentangled and highlight the class information to detect novelty class well. Low dimensional trait of latent space enable us to handle the features in the data easier. In this regard, we propose a novelty function using the change of angle in latent vectors by embedding input data in latent space orthogonal to each class using mutual class information.

\newcommand{\factorial}{\ensuremath{\mbox{\sc Factorial}}}

\section{Related work}
\textbf{One-class Novelty Detection.} In recent years, one-class novelty detection has received tremendous attentions as a traditional representation learning research problem. There have been many classical approaches to tackle this problem such as Principal Component Analysis (PCA). Deep learning, which has shown great performances in a variety of fields such as computer vision, cybersecurity, medical assistance, and etc., finds a way to learn representation and detect the based on previously seen representation. AE-based novelty detection mostly put reconstruction error such as mean squared error as a novelty detection function after learning the representation of the data. GAN-based novelty detection usually takes discriminator's prediction in the image space as a tool of measuring reconstruction error. One-Class novelty detection using GAN (OCGAN) shows a great performance in novelty detection in uni-modal normality data.

\noindent\textbf{Approaches on Novelty Score Function.} There has been other approaches to determine novelty scores other than reconstruction error or discriminator's prediction. Generative Probabilistic Novelty Detection with adversarial autoencoders (GPND)~\cite{noveltyaae} identifies novelty data by considering it to be an inlier or an outlier. GPND has done this by utilizing a probabilistic approach and computing how likely it is that a new data was generated by the normal distribution effectively. RaPP: Novelty Detection with Reconstruction along Projection Pathway (RaPP)\cite{rapp} introduces a new way to quantify novelty scores using values in hidden space activation obtained from a deep autoencoder. RaPP compares input and its autoencoder reconstruction both of in the input space and in all of the hidden spaces. However, in order to enforce their metrics, RaPP network is required to be symmetric, which makes designing network architecture and training network a very expensive work. As the data becomes more complicated, it becomes more expensive due to fully-connected layers in encoder and decoder caused by its structural problem. RaPP also showed a great performance in multi-modal normality case.

\section{Proposed Method: OAAE}
In this section, We propose a new AAE novelty detection algorithm using orthogonalized latent space (OAAE) for multi-modal normality case. The key idea is to disentangle latent space using mutual class information by employing orthogonal low rank embedding (OLE) loss\cite{ole}, which enables us to achieve minimizing the variance of latent vectors in intra-class as well as maximizing margins of inter-class latent vectors (in terms of angle; equivalently orthogonalize inter-class latent vectors). With such an orthogonalized latent space, we estimate a novelty score by quantifying the change of angle in each latent vector.

\subsection{Orthogonal Latent Embedding}
\begin{figure}[t] 
\centering
\includegraphics[width=\columnwidth]{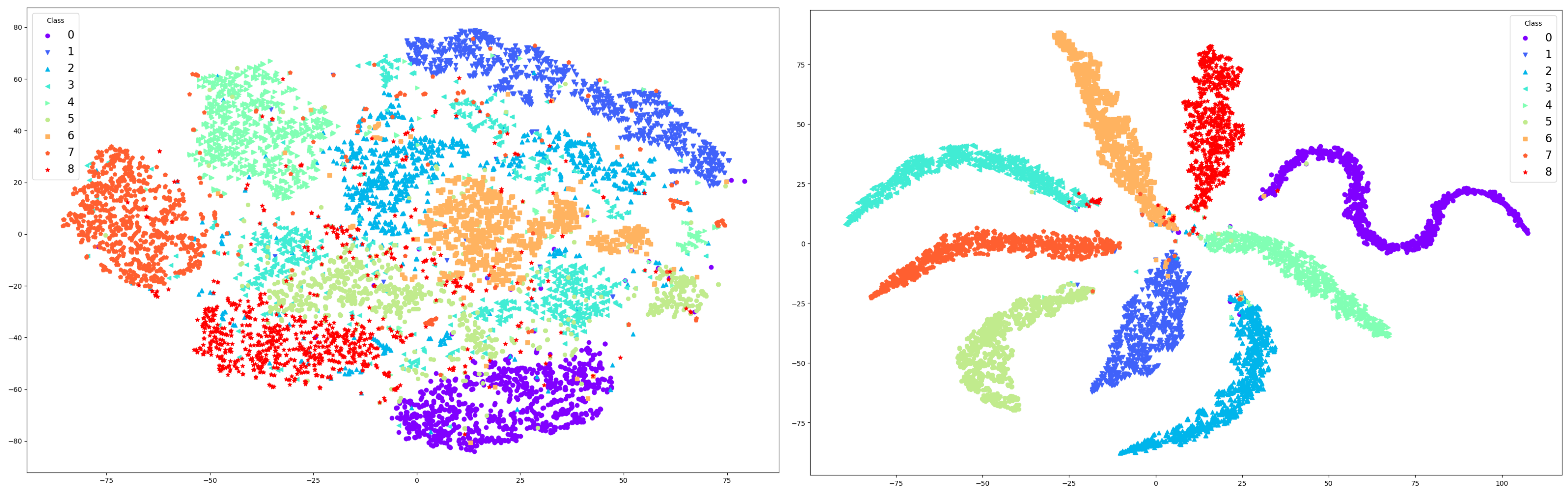}
\caption{Embedding of trained latent space using t-SNE. Left: AAE without OLE loss. Right: AAE with OLE loss. Reduced variance of intra-class clusters of latent vectors was observed.}
\label{oleffect}
\end{figure}

OLE is carried out using rank function~\cite{ole}. Mathematical formulations of OLE begins with the following equation:
\begin{equation} \label{eq:rank}
    \arg\min_{\mathbf{T}} \sum_{c=1}^{C}rank(\mathbf{TX}_c) - rank(\mathbf{TX}),\, \textrm{s.t.}\, ||\mathbf{T}||_2 = 1,
\end{equation}
where $\mathbf{X}$ denotes input dataset, $\mathbf{X}_c$ denotes the set of data points with class $c$ in a subspace of $\mathbf{R}^d$, $\mathbf{T}$ is a linear transformation on the data (i.e., feed forward network for deep learning framework), $||\cdot||_2$ is the matrix Euclidean norm. We interpret this formulation term by term intuitively~\cite{learningcluster}. Minimizing the first term $\sum_{c=1}^{C}rank(\mathbf{TX}_c)$ keeps the transformed data from the same subspace a consistent representation, and maximizing the second term $rank(\mathbf{TX})$ encourages the transformed data from different subspace to represent a diverse representation. Additionally, the normalization constraint $||\mathbf{T}||_2 = 1$ avoids the trivial solution, i.e., $\mathbf{T} = 0$. Since it is known that the nuclear norm ($||\textrm{\textbf{A}}||_\star$; the sum of the singular values of the matrix $\mathbf{A}$) is the convex envelop of $rank$(\textbf{A}) over the unit ball of matrices ~\cite{matrixrank}, and due to efficiency of optimization~\cite{robustpca,guaranteed}, we reformulate the equation using the nuclear norm  as follows:
\begin{equation} \label{eq:nun}
    \arg\min_{\mathbf{T}} \sum_{c=1}^{C} ||\mathbf{TX}_c||_\star - ||\mathbf{TX}||_\star,\, \textrm{s.t.}\, ||\mathbf{T}||_2 = 1.
\end{equation}
Following \cite{ole}, (\ref{eq:nun}) becomes the following loss using minibatch as below to be applied to the deep learning framework:
\begin{align}\label{eqn:nndl}
    \mathbf{L}_{OLE}(\mathbf{Y})&:=\sum_{c=1}^C \max(\Delta, ||\mathbf{Y}_c||_\star) - ||\mathbf{Y}||_\star\\
    &= \sum_{c=1}^C \max(\Delta, ||\Phi(\mathbf{X}_c;\theta)||_\star) - ||\Phi(\mathbf{X};\theta)||_\star.
\end{align}
To optimize (\ref{eqn:nndl}) using backpropagation, the projected subgradient for the nuclear norm and the descent direction for (\ref{eqn:nndl}) are obtained in  by using SVD decomposition on matrix $\mathbf{A}$, i.e., $\mathbf{A = U\Sigma V}^T$, and zero filling matrix $\mathbf{Z}_c$ as follows:
\begin{equation}\label{subgradnn}
    g_{||A||_\star}(A) = \mathbf{U}_1 \mathbf{V}_1^T,
\end{equation}
\begin{equation}\label{lossgd}
    g_{\mathbf{L}_{OLE}}(\mathbf{Y})=\sum_{c=1}^C\left[ \mathbf{Z}_c^{(l)} |\mathbf{U}_{c1} \mathbf{U}_{c1}^{T} | \mathbf{Z}_c^{(r)} \right] - \mathbf{U}_1\mathbf{V}_1^{T}.
\end{equation}
where $\mathbf{U}_1$ and $\mathbf{V}_1$ be the first $s$ columns of $\mathbf{U}$ and $\mathbf{V}$, respectively, corresponding to eigenvalues larger than a small threshold value $\delta$. Similarly, $\mathbf{U}_{c1}$ and ${\mathbf{V}_{c1}}$ be left and right singular vectors of $\mathbf{Y}_c$ where their corresponding singular values are greater than the threshold $\delta$. Using $\mathbf{L}_{OLE}$ loss, we embed our high dimension dataset in orthgonoalized latent space with the two main benefits: reduced variance of intra-class, maximized angle margins of clusters of inter-class as shown in Figure \ref{oleffect}. 

\subsection{Architecture}
\begin{figure}[t] 
\centering
\includegraphics[width=\columnwidth]{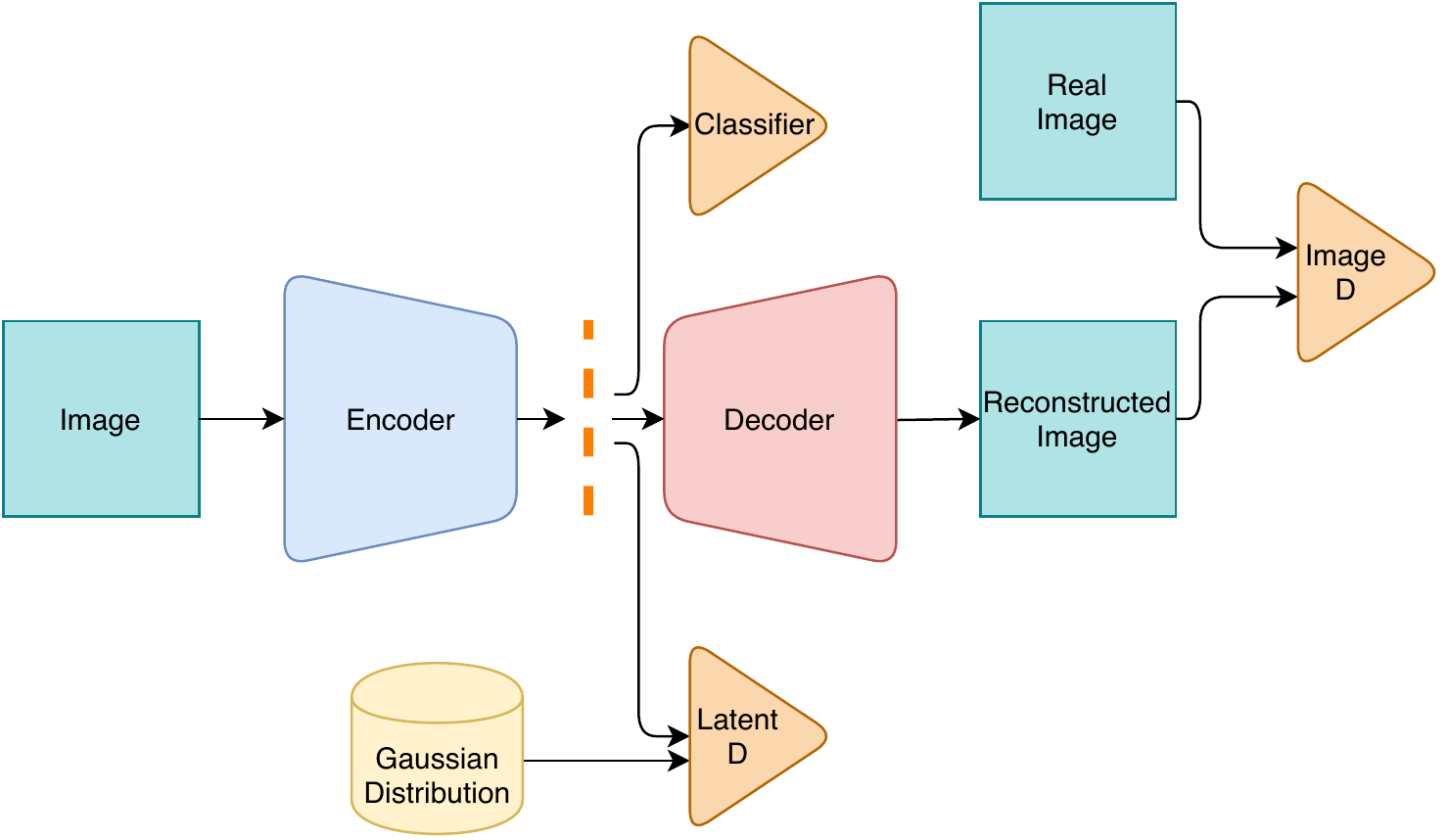}
\caption{OAAE architecture}
\label{archi}
\end{figure}

The architecture of the proposed network is based on AAE ~\cite{aae} and classifier was added to use mutual class information in OLE loss shown in Figure \ref{archi}. Each of encoder and decoder in our model has five layers with three convolutional layers and two fully-connected layers at the end. Details of training of our algorithm is described in Algorithm\ref{algorithm}. Main key in our algorithm is to adopt OLE loss to use mutual information and disentangle features in latent space and returns novelty score using the change of angles in latent vectors. 

\begin{algorithm}
\caption{Novelty Detection algorithm}\label{euclid}
\begin{algorithmic}[1]
\State \textbf{Input} : Image $x$ with class $c$, $N$ Epochs, $K$ Iteration
\State \textbf{Training phase}
\For{epochs $0$ to $N$}
    \For{iteration $0$ to $K$}
    \State $n \gets \mathcal{N}(0,I)$
    \State $z \gets \mathcal{N}(0,I)$
    \State Discriminator training phase
    \State $\mathcal{L}_{latent} \gets \mathcal{D}_{latent}(z,1) + \mathcal{D}_{latent}(Enc(x+n),0)$
    \State $\mathcal{L}_{image} \gets \mathcal{D}_{image}(x,1) + \mathcal{D}_{image}(Dec(z),0)$
    \State Back-propagate and update
    \State Encoder, Decoder and Classifier training phase
    \If{K\%5 == 0}
    \State $\mathcal{L}_{recon} \gets ||{x - Dec(Enc(x+n))}||^2_2 $
    \State $\mathcal{L}_{Enc} \gets \mathcal{D}_{latent}(Enc(x+n),1)$
    \State $\mathcal{L}_{Dec} \gets \mathcal{D}_{image}(Dec(z),1)$
    \State $\mathcal{L}_{ole} \gets OLE(Enc(x+n),c)$
    \State $\mathcal{L}_{cls} \gets CrossEntropy(C(Enc(x+n)),c)$
    \State Back-propagate and update
    \EndIf
    
    \EndFor
\EndFor

\State \textbf{Test phase}
\State Test image $x$
\State $z_0 \gets Enc(x)$
\State $z_1 \gets Enc(Dec(Enc(x)))$
\State $Novelty\_Score \gets angle(z_0,z_1)$

\end{algorithmic}
\label{algorithm}
\end{algorithm}

\section{Experiment}

\begin{table*}[t]
\centering
\caption{AUROC of OAAE and the baselines}
\smallskip
\begin{tabular}{llllllllllll}
\hline
& & & & &MNIST\\ \hline\hline
& 0 & 1 & 2 & 3 & 4 & 5 & 6 & 7 & 8 & 9 & Mean\\ \hline
 OCGAN & 0.91 &0.08&0.76&0.81&0.77&0.72&0.87&0.37&0.923&0.46&0.67\\
 RaPP &  \textbf{0.99}&0.89&\textbf{0.98}&\textbf{0.95}&0.92&\textbf{0.97}&\textbf{0.98}&0.97&0.96&0.89&0.95\\
 OAAE &  0.98 &\textbf{0.97}&0.97&\textbf{0.95}&\textbf{0.95}&
 \textbf{0.97}&0.975&\textbf{0.972}&\textbf{0.982}&\textbf{0.968}&\textbf{0.970}\\ \hline \\ 
& & & & &f-MNIST\\ \hline\hline 
 & T-shirt &Trouser&Pullover&Dress&Coat&Sandals&Shirt&Sneaker&Bag&Boots&Mean\\ \hline
OCGAN&0.577&0.750&0.596&0.723&0.557&0.801&0.546&0.769&0.877&0.726&0.692\\
RaPP&0.70&0.78&0.65&0.82&0.57&\textbf{0.85}&0.58&0.61&\textbf{0.98}&\textbf{0.82}&0.736\\ OAAE&\textbf{0.915}&\textbf{0.88}&\textbf{0.816}&\textbf{0.847}&\textbf{0.853}&0.716&\textbf{0.791}&\textbf{0.789}&0.966&0.799&\textbf{0.837}\\ \hline \\ 
& & & & & CIFAR10\\ \hline\hline
&  Airplane&Automobile&Bird&Cat&Deer&Dog&Frog&Horse&Ship&Truck&Mean\\ \hline

OCGAN&0.54&0.71&0.40&0.52&0.31&0.58&0.40&0.61&0.44&0.69&0.52\\
RAPP&0.469&0.654&0.416&0.578&0.357&0.604&0.382&0.579&0.553&0.681&0.527\\
OAAE&\textbf{0.706}&\textbf{0.777}&\textbf{0.579}&\textbf{0.713}&\textbf{0.660}&\textbf{0.742}&\textbf{0.620}&\textbf{0.683}&\textbf{0.652}&\textbf{0.786}&\textbf{0.692}\\ \hline
\end{tabular}
\label{table1}
\end{table*}

\subsection{Datasets}
\textbf{MNIST}. The MNIST database, which stands for Modified National Institute of Standards and Technology database, consists of a large number of 28$\times$28 gray scale images of handwritten digits (10 classes; 0$\sim$9). The MNIST dataset is commonly and widely used for various computer vision, image processing researches due to its simplicity. In our experiments, we choose images of one handwritten digit and every other images of remaining nine different handwritten digits as a novelty class, normal class data, respectively.

\noindent\textbf{Fasion MNIST (f-MNIST)}. The fashion-MNIST is a dataset of 28$\times$28 grayscale images 10 different classes (T-shirt, Trouser, Pullover, Dress, Coat, Sandals, Shirt, Sneaker, Bag, Ankle boots). It shares the same image size with the original MNIST dataset but f-MNIST is regarded as a harder data to learn in general because of the complexity that semantic images have. Similar to the previous experiments on MNIST dataset, we choose images with one class (e.g., T-shirt) and every other images of remaining nine different class as a novelty class, normal class data, respectively.

\noindent\textbf{CIFAR10}. The CIFAR10 dataset consists of 60000 32$\times$32 coloured images with evenly distributed 10 classes (airplane, automobile, bird, cat, deer, dog, frog, horse, ship, truck). This dataset was selected due to its complexity. CIFAR10 dataset is usually treated as harder data to train than MNIST or f-MNIST in general due to its multi-channel trait. 

\subsection{Architectures of Baseline Algorithms}
We compare performance of our models to that of two state-of-the-art GANs-based novelty detection algorithms: OCGAN, RaPP. We briefly explain how those two algorithms work in the following sections.

\noindent\textbf{OCGAN}. OCGAN solves classical one-class novelty detection problem and aims to determine whether the new input is from the same class or not. The key idea of OCGAN is to learn latent representations of normal class data using a denoising autoencoder network and to directly force the latent space to entirely represent the given class. OCGAN is particularly focused on learning uni-modal normality data.
%In order to do that, OCGAN force the latent space to have bounded support by introducing a tanh activation in the encoder’s output layer. Secondly, using a discriminator in the latent space that is trained adversarially, OCGAN ensure that encoded representations of in-class examples resemble uniform random samples drawn from the same bounded space. Thirdly, using a second adversarial discriminator in the input space, OCGAN ensure all randomly drawn latent samples generate examples that look real. Finally, we introduce a gradient-descent based sampling technique that explores points in the latent space that generate potential out-of-class examples, which are fed back to the network to further train it to generate in-class examples from those points. 

\noindent\textbf{RaPP}. A new methodolgy for novelty detection is proposed in RaPP by adopting values in hidden space activation obtained from a deep AE. RAPP compares input and its AE or VAE reconstruction in the hidden spaces as well as in the input space. RaPP introduces two metrics combining those hidden activated values to measure novelty scores. In order to achieve this, RaPP requires the model to be symmetric to enforce its evaluation methodologies, which causes its structural limitation, and training model becomes a very expensive work as the data becomes more complicated due to fully-connected layers in encoder and decoder caused by their structural problem.
\subsection{Training Details}
All of our experiments were conducted by Python 3.6.9. Adam optimizer was adopted to train our model. For the stable adversarial learning, the encoder is trained with one iteration after every five iterations for the discriminator. Each experiment is carried out with 100 epochs with batch size as much as 64, and we set learning rate as 0.0004. Gaussian noises with standard deviation of 0.02 were added to the input image data at the training phase.
\subsection{Experimental Results}
We evaluate the performances of all experiments using Area Under the Receiver Operating Characteristic curve (AUROC) as shown in Table \ref{table1}.
\section{Discussion}
Our methods showed a better performance than other previous GAN-based state-of-the-art novelty detection algorithms such as OCGAN, RaPP. Specifically, our approach provides a much higher AUROC values for experiments on more complicated data such as f-MNIST, CIFAR-10. It supports that as a tool of novelty score measurement, change of latent vector is more reasonable than image reconstruction errors since image reconstruction error can be more escalated in more complicated data. In training level, our approach leverages on class labels in normal dataset, which is sometimes a expensive work. Unsupervised learning framework without using normal class labels can be considered potentially.

\section{Conclusion}
We proposed a new novelty detection framework using deep generative models. Instead of evaluating novelty class using image reconstruction error, the change of angle in latent vector is regarded as a tool for novelty detection quantity. We adopt OLE loss using mutual class information to achieve disentanglement of latent vectors to maximize the effect of class information. Our new approach shows a greater performance in multi-modal normality scenarios than previously existing GAN based state-of-the-art novelty detection algorithms.
\bibliographystyle{aaai}
\bibliography{main.bib}

\end{document}